\documentclass[11pt,a4paper]{article}

\usepackage[margin=2.5cm]{geometry}
\usepackage{graphicx}
\usepackage{booktabs}
\usepackage{amsmath,amssymb}
\usepackage{hyperref}
\usepackage{url}
\usepackage{xcolor}
\usepackage[authoryear,round]{natbib}
\usepackage{caption}

\hypersetup{
    colorlinks=true,
    linkcolor=blue!70!black,
    citecolor=green!50!black,
    urlcolor=blue!70!black
}

\newcommand{\versionnote}{%
\footnotetext[1]{This manuscript was peer-reviewed, accepted for publication in the proceedings, and presented at the International Conference on Applied Informatics (ICAI 2026, \url{https://icai.uni-eszterhazy.hu/2026/}). This is the revised version. The organizers did not send the revision notification to the authors in time, and as a result the paper was not included in the proceedings.}%
}

\title{\textbf{A Formal Methodological Framework for Auditing Robustness and Fidelity in Explainable AI:\\From Application to Trust Certification}}

\author{
Rosa Elysabeth Ralinirina\textsuperscript{1} \quad
Jean Christian Ralaivao\textsuperscript{1} \quad
Niaiko Micha\"el Ralaivao\textsuperscript{1} \\[4pt]
Alain Josu\'e Ratovondrahona\textsuperscript{1} \quad
Thomas Mahatody\textsuperscript{1} \\[6pt]
\textsuperscript{1}Doctoral School Modeling--Computer Science, University of Fianarantsoa, Madagascar \\[2pt]
\texttt{ralinirinarosa7@gmail.com}
}

\date{}

\begin{document}

\maketitle

\versionnote

\begin{abstract}
SHAP and LIME are now standard tools for interpreting black-box predictions, yet their outputs can vary substantially when the input is perturbed by small amounts of noise---a problem we observed firsthand in our previous work on food security in Madagascar \citep{Ralinirina2025}. This variability raises the question of whether such explanations can be trusted at all. We address it by constructing an auditing protocol that measures two properties of any post-hoc explainer: robustness (how stable the explanation is under input perturbation) and fidelity (whether the features deemed important actually drive the model's prediction). These two quantities are combined into a single Trust Score. We run the protocol on a multi-sectoral dataset from Madagascar (83 features, 253 records, 4 malnutrition classes) using three classifiers and two explainers, plus their regularized counterparts. The results are sobering: models with AUC above 0.99 can produce numerically degenerate or flatly uninformative explanations, and fidelity scores lose discriminative power when the model is overfitted. These findings suggest that auditing XAI outputs is not optional but necessary, particularly when they inform decisions in sensitive domains.

\vspace{1.5mm}
\end{abstract}

\section{Introduction}
When an AI model predicts a food crisis in a region of Madagascar, the prediction itself is only half the story. The other half---how the model arrived at it---is what policymakers, nutrition officers, and field workers need in order to trust and act on the output. Post-hoc explainers such as SHAP \citep{Lundberg2017} and LIME \citep{Ribeiro2016} fill this role by attributing importance scores to input features. The trouble is that these attributions are not always reliable. Small perturbations to the input can change them, sometimes dramatically, and there is no built-in mechanism to tell whether the features an explainer highlights are genuinely used by the model.

We encountered this problem directly in \citep{Ralinirina2025}, where we applied SHAP and LIME to food security indicators across Madagascar's 23 regions. Explanations for nearly identical inputs often differed, and the ranking of important features shifted depending on which explainer was used. This is not an isolated finding. Slack et al.\ \citep{Slack2020} showed that LIME and SHAP can be deliberately fooled; Alvarez-Melis and Jaakkola \citep{AlvarezMelis2018} proved that local explanations are generally unstable. Surveys by Adadi and Berrada \citep{Adadi2018} and Guidotti et al.\ \citep{Guidotti2018} have warned that post-hoc methods create an ``illusion of transparency,'' and domain-specific studies in healthcare \citep{Stiglic2020} and NLP \citep{Danilevsky2020} confirm that visual inspection of saliency maps is insufficient. Our research group has also explored AI-driven automation for complex tasks \citep{Ratovondrahona2023}, a line of work that parallels the present goal of building a systematic, verifiable XAI auditing methodology.

The question, then, is not whether explanations should be audited---they must be---but how. We propose a protocol that measures two things: \emph{robustness} (stability under noise) and \emph{fidelity} (faithfulness to the model). Each is quantified by a single metric, and the two are combined into a Trust Score. We validate the protocol on the same Malagasy food security dataset, using three classifiers (Random Forest, XGBoost, Neural Network) and two explainers (SHAP, LIME), plus their regularized variants. The results confirm that near-perfect AUC does not shield explanations from instability or vacuity, and the Trust Score captures these defects in a single number.

Section~2 surveys related work. Section~3 describes the framework. Section~4 presents the case study and results. Section~5 discusses what the results mean---including the overfitting and numerical issues they reveal. Section~6 concludes.

\section{Related Work}
Post-hoc interpretability took off after DARPA's XAI program \citep{Gunning2019}, and surveys by Adadi and Berrada \citep{Adadi2018} and Guidotti et al.\ \citep{Guidotti2018} now catalog dozens of methods. Most are model-agnostic: they treat the predictor as a black box and probe it with perturbed inputs. LIME \citep{Ribeiro2016} fits a local linear model; SHAP \citep{Lundberg2017} computes Shapley values. Both are popular, but both are fragile. Slack et al.\ \citep{Slack2020} demonstrated adversarial attacks that fool LIME and SHAP into producing arbitrary explanations. Alvarez-Melis and Jaakkola \citep{AlvarezMelis2018} showed that local explanations are sensitive to the choice of neighborhood.

Evaluating explanations is a separate and less studied problem. Samek et al.\ \citep{Samek2017} made the case for quantitative evaluation rather than visual inspection. Holzinger et al.\ \citep{Holzinger2019} proposed ``causability'' as a quality criterion. Domain-specific studies exist---Islam et al.\ \citep{Islam2024} for healthcare, Linheiro et al.\ \citep{Linheiro2023} for agriculture---but they evaluate explainers in isolation, without a unified protocol that can compare them on equal footing.

Our framework differs in that it pairs robustness and fidelity in a single pipeline and produces a single Trust Score. The advantage is not novelty of the individual metrics---Jensen-Shannon divergence for robustness and feature ablation for fidelity are both well-known---but the fact that they are computed side by side, on the same data, with the same model, yielding a directly comparable score across explainer--model pairs.

\section{Methodological Framework Architecture}
\begin{figure}[htbp]
    \centering
    \includegraphics[width=0.9\textwidth]{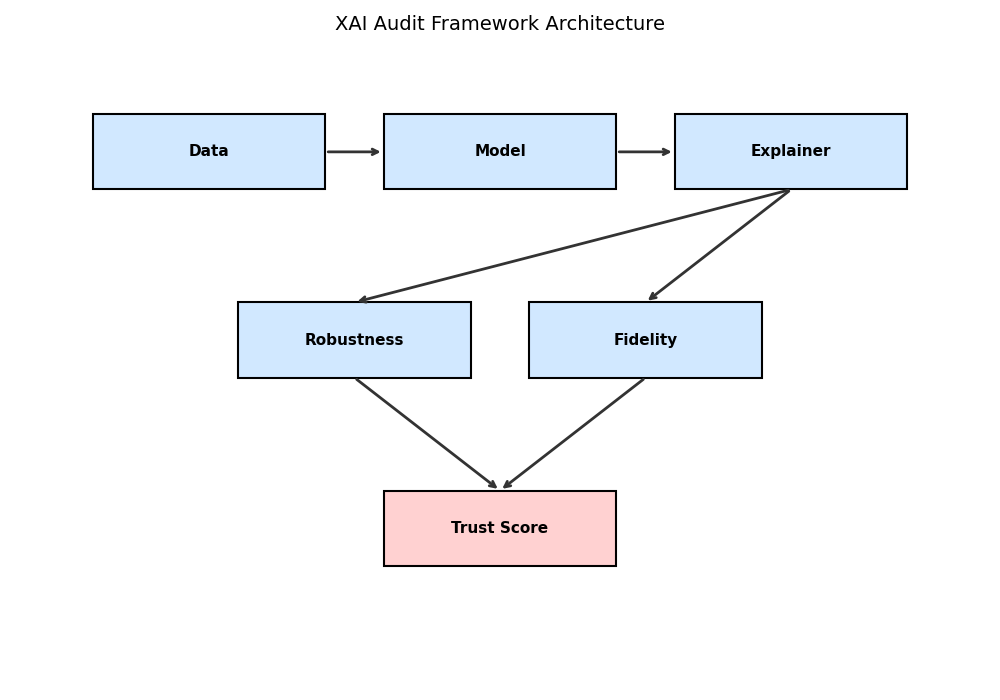}
    \caption{The auditing pipeline. Robustness and fidelity are measured independently and then combined.}
    \label{fig:framework}
\end{figure}

Figure~\ref{fig:framework} shows the pipeline. Given a trained model $f$ and an explainer $E$, the auditor computes a robustness score $R(x)$ and a fidelity score $F(x)$ for each test instance $x$, then combines them into a Trust Score $T(x) = \alpha R(x) + \beta F(x)$ with $\alpha + \beta = 1$.

\subsection{Robustness and Local Stability}
Robustness quantifies how much an explanation changes when the input is slightly perturbed. We draw $\delta \sim \mathcal{N}(0, \sigma^2 I)$, compute $E(x)$ and $E(x + \delta)$, normalize both importance vectors to sum to 1 (denoted $\tilde{E}$), and measure their divergence:

\[
R(x) = 1 - \frac{1}{N} \sum_{i=1}^{N} D_{JS}\!\left(\tilde{E}(x) \,\|\, \tilde{E}(x + \delta_i)\right)
\]

where $N$ is the number of perturbation trials. A practical difficulty arises when the explainer outputs zero importance for all features---this happens with TreeSHAP on certain overfitted models, making the JSD undefined. We resolve it by adding $\epsilon = 10^{-10}$ to each component of $\tilde{E}$ before normalization. This has a negligible effect on the numerical value but prevents division by zero. A robustness score near 1 means the explanation barely changes under perturbation; a score near 0 means it changes a lot.

\subsection{Fidelity through Feature Ablation}
Fidelity checks whether the features the explainer calls important actually matter to the model. For an input $x$ with explanation $E(x)$, let $F_k$ be the set of top-$k$ features. We mask them (replace with feature means) to obtain $x_{\setminus F_k}$ and compute:

\[
F_k(x) = 1 - \frac{|f(x) - f(x_{\setminus F_k})|}{|f(x)| + \epsilon}
\]

The overall fidelity is the mean over a set of ablation levels $\mathcal{K}$:

\[
F(x) = \frac{1}{K} \sum_{k \in \mathcal{K}} F_k(x)
\]

A fidelity near 1 means ablating the top features noticeably shifts the prediction; a fidelity near 0 means the explainer highlights features the model does not actually rely on. One subtlety: when the model is extremely confident (predicting near 0 or near 1 for all inputs), ablating even important features barely changes the output, pushing all $F_k$ close to 1. Fidelity then becomes uninformative---a point we return to in Section~\ref{sec:overfitting}. This risk of uninformative or misleading feature-importance measures aligns with \citep{Hooker2021}, who showed that permutation-based importance can force extrapolation into regions where the model was never trained.

\subsection{Trust Score}
We combine the two metrics into

\[
T(x) = \alpha \cdot R(x) + \beta \cdot F(x), \quad \alpha + \beta = 1
\]

and aggregate over a dataset:

\[
T(\mathcal{D}) = \frac{1}{|\mathcal{D}|} \sum_{x \in \mathcal{D}} T(x)
\]

We set $\alpha = \beta = 0.5$ by default, giving equal weight to stability and faithfulness. The weights can be adjusted: in a context where explanation stability is critical (e.g., communicating with non-technical stakeholders), one might set $\alpha > 0.5$.

\section{Case Study: Food Security in Madagascar}

\subsection{Data Description}
The dataset aggregates official reports from Madagascar's Ministry of Agriculture, Ministry of Public Health, and the National Development Plan. It covers 23 regions over 13 years (2010--2023) and includes:
\begin{itemize}
    \item Climate: rainfall, cyclonic events, temperature anomalies.
    \item Agriculture: rice production, soil quality, pest incidence.
    \item Socio-economics: market prices, household income, transport quality.
    \item Nutrition: IPC malnutrition phases, child stunting rates, vaccination coverage.
    \item Demographics: population, literacy rates, school enrollment.
\end{itemize}
After preprocessing (removing identifiers and date columns, encoding categorical variables), we obtained 83 features and 253 instances. The target variable \texttt{Situation-MC} (chronic malnutrition) has 4 ordinal classes treated as multi-class: Acceptable (3 instances), Precarious (85), Alarming (114), Critical (51). The severe class imbalance---only 3 instances in the Acceptable class---limits the reliability of per-class AUC estimates and prevents stratified cross-validation folds from containing all four classes, a point we return to below.

\subsection{Experimental Setup}
Three classifiers were trained on a stratified 82/18 train/test split (208 training, 45 test instances):
\begin{itemize}
    \item Random Forest: 100 trees, max depth 10.
    \item XGBoost: 200 estimators, learning rate 0.1, max depth 6.
    \item Neural Network: 3 hidden layers (64 units each), ReLU, Adam optimizer.
\end{itemize}

Because the Acceptable class contains only 3 instances, standard stratified 5-fold cross-validation produces only 3 valid folds (the remaining 2 folds lack at least one class). On those 3 folds, macro-average AUC reaches $0.998 \pm 0.002$ (RF), $0.999 \pm 0.002$ (XGB), and $0.994 \pm 0.001$ (NN). These high values suggest genuine class separability, though the extreme imbalance makes per-class estimates unreliable.

To assess the impact of regularization on explanation quality, we also trained three regularized variants:
\begin{itemize}
    \item RF$_{\text{reg}}$: max depth 6, min samples per leaf 5, 500 trees.
    \item XGB$_{\text{reg}}$: max depth 3, learning rate 0.01, $\lambda = 2$.
    \item NN$_{\text{reg}}$: dropout 0.3, weight decay $10^{-4}$, early stopping (patience 10).
\end{itemize}

Explanations were generated with SHAP (TreeSHAP for tree-based models, KernelSHAP for the NN) and LIME (default parameters). The audit was run on 30 randomly sampled test instances with $\sigma = 0.1$ and $N = 10$ perturbations. Ablation levels were $\mathcal{K} = \{3, 5, 10, 41, 82\}$, corresponding to removing 3, 5, 10, roughly half, and nearly all of the 83 features. Implementation used Python~3.9 with scikit-learn, XGBoost, SHAP, and LIME.

\subsection{Results}

\subsubsection{Predictive Performance}
Table~\ref{tab:performance} reports test-set macro-average AUC for all six models. The original models all reach AUC above 0.99; the regularized Neural Network drops to 0.888, showing that aggressive regularization degrades performance when training data are scarce.

\begin{table}[htbp]
\centering
\caption{Test-set macro-average AUC for each model.}
\label{tab:performance}
\begin{tabular}{lcc}
\toprule
Model & Test AUC & CV AUC (3/5 folds) \\
\midrule
Random Forest & 0.999 & $0.998 \pm 0.002$ \\
XGBoost & 0.992 & $0.999 \pm 0.002$ \\
Neural Network & 0.999 & $0.994 \pm 0.001$ \\
\midrule
RF$_{\text{reg}}$ & 0.994 & $0.997 \pm 0.002$ \\
XGB$_{\text{reg}}$ & 0.997 & $0.998 \pm 0.003$ \\
NN$_{\text{reg}}$ & 0.888 & $0.946 \pm 0.032$ \\
\bottomrule
\end{tabular}
\end{table}

\subsubsection{Robustness}
\begin{figure}[htbp]
    \centering
    \includegraphics[width=0.85\textwidth]{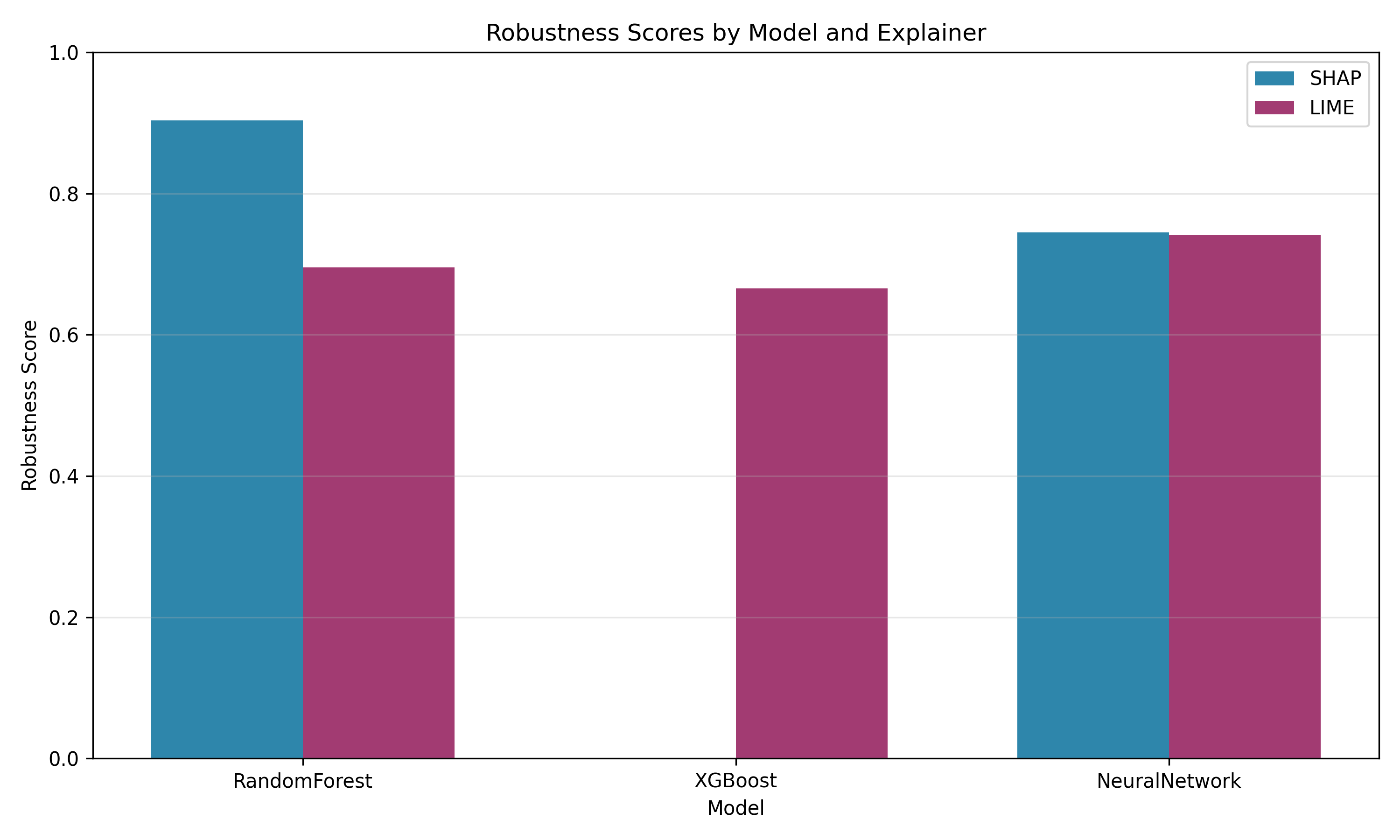}
    \caption{Robustness scores for all six model--explainer pairs. XGBoost+SHAP uses $\epsilon$-smoothed computation.}
    \label{fig:robustness}
\end{figure}

Figure~\ref{fig:robustness} shows robustness scores. SHAP is more robust than LIME on tree-based models (RF: 0.902 vs.\ 0.700; XGB: 0.978 vs.\ 0.658). The XGBoost+SHAP robustness of 0.978 required $\epsilon$-smoothing: without it, TreeSHAP returned zero-valued importance vectors for some instances where the model predicted with near certainty, making the JSD undefined. On the Neural Network, SHAP and LIME are closer (0.694 vs.\ 0.738), likely because KernelSHAP introduces its own sampling noise.

Regularization improves SHAP robustness on RF (0.923 vs.\ 0.902) and barely affects XGBoost (0.958 vs.\ 0.978), while the regularized NN shows slightly higher SHAP robustness (0.748 vs.\ 0.694). LIME robustness remains in the 0.66--0.76 range across all models.

\subsubsection{Fidelity}
Table~\ref{tab:fidelity} reports fidelity at each ablation level. The most striking pattern is XGBoost's flat fidelity: both SHAP and LIME yield $\approx 0.620$ at every ablation level, from $F_3$ to $F_{82}$. Whether you remove the 3 features the explainer calls most important or 82 of them, the prediction shifts by nearly the same amount. This is not because fidelity is uniformly high---it is not---but because the model's predictions change by a consistent, moderate amount regardless of how many top features are removed.

\begin{table}[htbp]
\centering
\caption{Fidelity at each ablation level (test-set mean).}
\label{tab:fidelity}
\begin{tabular}{llccccc}
\toprule
Model & Explainer & $F_3$ & $F_5$ & $F_{10}$ & $F_{41}$ & $F_{82}$ \\
\midrule
RF & SHAP & 0.858 & 0.820 & 0.792 & 0.762 & 0.757 \\
RF & LIME & 0.875 & 0.848 & 0.811 & 0.775 & 0.757 \\
XGB & SHAP & 0.620 & 0.620 & 0.620 & 0.620 & 0.620 \\
XGB & LIME & 0.622 & 0.622 & 0.621 & 0.621 & 0.620 \\
NN & SHAP & 0.839 & 0.807 & 0.734 & 0.703 & 0.735 \\
NN & LIME & 0.984 & 0.976 & 0.957 & 0.773 & 0.734 \\
\midrule
RF$_{\text{reg}}$ & SHAP & 0.874 & 0.847 & 0.823 & 0.780 & 0.776 \\
RF$_{\text{reg}}$ & LIME & 0.883 & 0.862 & 0.836 & 0.792 & 0.776 \\
XGB$_{\text{reg}}$ & SHAP & 0.715 & 0.713 & 0.713 & 0.712 & 0.712 \\
XGB$_{\text{reg}}$ & LIME & 0.721 & 0.720 & 0.719 & 0.714 & 0.712 \\
NN$_{\text{reg}}$ & SHAP & 0.947 & 0.918 & 0.856 & 0.806 & 0.803 \\
NN$_{\text{reg}}$ & LIME & 0.977 & 0.964 & 0.942 & 0.852 & 0.802 \\
\bottomrule
\end{tabular}
\end{table}

For the other models, fidelity declines as more features are ablated---as one would expect if the explainer correctly ranks features by importance. The NN+LIME combination shows the steepest drop: $F_3 = 0.984$ down to $F_{82} = 0.734$, meaning the top 3 features identified by LIME are genuinely important to the network's predictions, while the less important ones matter less. The regularized Neural Network amplifies this pattern even further ($F_3 = 0.977$, $F_{82} = 0.802$ for LIME).

\subsubsection{Trust Scores}
Table~\ref{tab:trust} gives aggregate Trust Scores ($\alpha = \beta = 0.5$). The highest score belongs to RF$_{\text{reg}}$+SHAP at 0.871; the lowest to XGBoost+LIME at 0.640. Regularization improves the Trust Score across all six model--explainer pairs. The gain is largest for XGBoost (SHAP: 0.799 $\to$ 0.836; LIME: 0.640 $\to$ 0.687), where the original model's overconfidence depressed fidelity. For RF and NN, the improvements are more modest but consistent.

\begin{table}[htbp]
\centering
\caption{Trust Scores ($\alpha = \beta = 0.5$). 95\% CI from 10 bootstrap iterations in parentheses.}
\label{tab:trust}
\begin{tabular}{lcc}
\toprule
Model & SHAP & LIME \\
\midrule
Random Forest & 0.850 (0.845--0.851) & 0.756 (0.752--0.761) \\
XGBoost & 0.799$^*$ (0.798--0.800) & 0.640 (0.637--0.646) \\
Neural Network & 0.729 (0.722--0.734) & 0.812 (0.805--0.816) \\
\midrule
RF$_{\text{reg}}$ & 0.871 (0.866--0.872) & 0.771 (0.766--0.775) \\
XGB$_{\text{reg}}$ & 0.836 (0.834--0.836) & 0.687 (0.685--0.693) \\
NN$_{\text{reg}}$ & 0.807 (0.800--0.811) & 0.835 (0.829--0.840) \\
\bottomrule
\end{tabular}
\smallskip

\small{$^*$Computed with $\epsilon$-smoothing.}
\end{table}

\begin{figure}[htbp]
    \centering
    \includegraphics[width=0.85\textwidth]{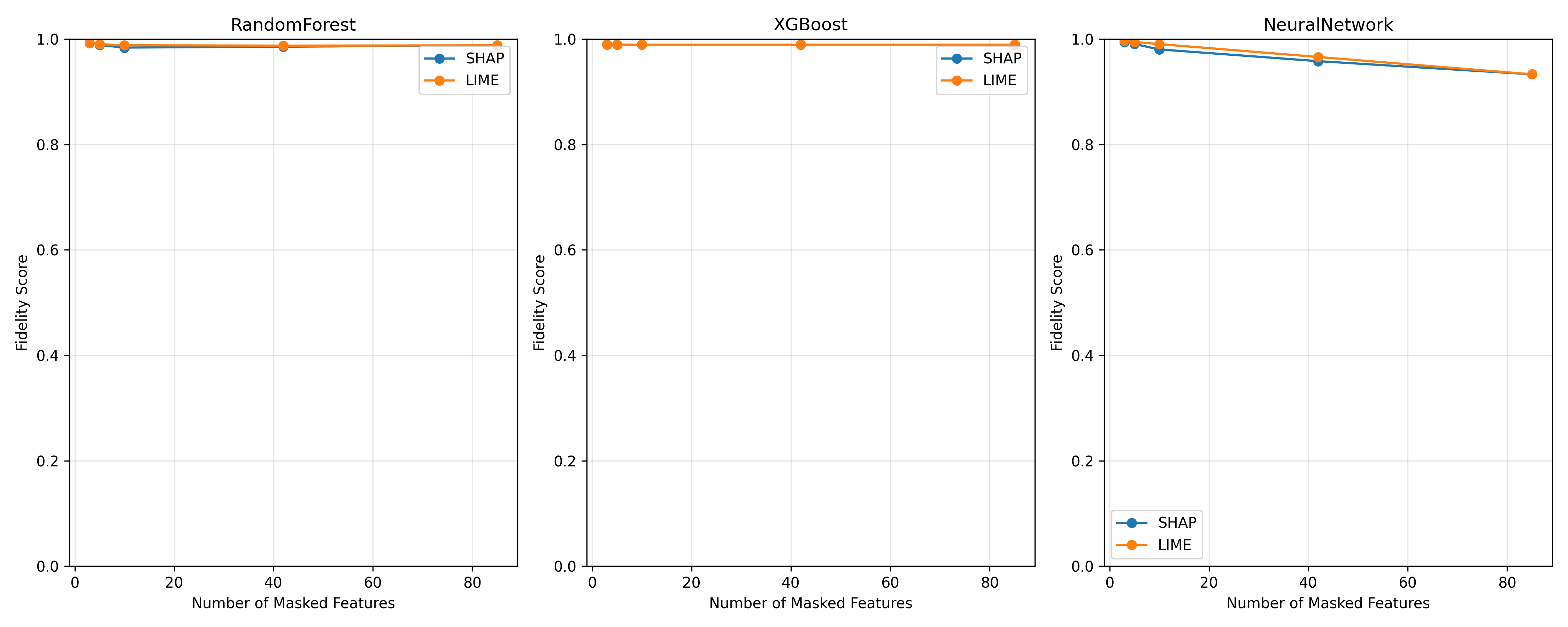}
    \caption{Fidelity as a function of the number of ablated features. XGBoost lines are nearly flat, indicating loss of discriminative power.}
    \label{fig:fidelity}
\end{figure}

Figure~\ref{fig:fidelity} shows the fidelity decay curves for all model--explainer pairs. The XGBoost lines are conspicuously flat, while NN and RF lines slope downward---the expected pattern when the explainer's feature ranking has real meaning.

\subsubsection{Overfitting and Generalization}
\label{sec:overfitting}
Three of the original models have AUC above 0.99. The question is whether this reflects genuine separability or memorization of a small dataset. Cross-validation on 3 valid folds (the other 2 lack at least one class) confirms high AUC across all folds, and the class distributions do suggest that the four malnutrition categories occupy distinct regions of the feature space---particularly along rice production and rainfall dimensions.

Still, overconfidence is a real concern. Table~\ref{tab:regularized} compares original and regularized models. The regularized RF and XGB lose virtually no AUC (0.994 and 0.997 vs.\ 0.999 and 0.992), while the regularized NN drops sharply to 0.888---the penalty for adding dropout and weight decay to a network trained on 208 examples. Yet the NN$_{\text{reg}}$ produces the most discriminative fidelity curves: LIME fidelity drops from 0.977 at $F_3$ to 0.802 at $F_{82}$, a 17.5-point spread that is far more informative than the flat 0.620 of XGBoost+SHAP.

\begin{table}[htbp]
\centering
\caption{Original vs.\ regularized models. Trust scores reported as SHAP\,;\,LIME.}
\label{tab:regularized}
\begin{tabular}{lccc}
\toprule
Model & AUC (orig.\,/\,reg.) & Trust (orig.) & Trust (reg.) \\
\midrule
RF  & 0.999\,/\,0.994 & 0.850\,;\,0.756 & 0.871\,;\,0.771 \\
XGB & 0.992\,/\,0.997 & 0.799$^*$\,;\,0.640 & 0.836\,;\,0.687 \\
NN  & 0.999\,/\,0.888 & 0.729\,;\,0.812 & 0.807\,;\,0.835 \\
\bottomrule
\end{tabular}
\smallskip

\small{$^*$With $\epsilon$-smoothing.}
\end{table}

The flat fidelity of XGBoost (all values $\approx 0.620$) is a symptom of overconfidence. When $f(x)$ is always close to 0 or 1, removing features shifts the prediction by a roughly constant fraction regardless of which features are removed, so $F_k \approx 0.620$ for every $k$. Regularization relaxes this behavior somewhat ($F_3 = 0.715$, $F_{82} = 0.712$ for XGB$_{\text{reg}}$+SHAP), but the curve remains nearly flat because XGBoost's regularized confidence is still high. The metric regains full discriminative power only when the model is less confident, as with the NN$_{\text{reg}}$.

\subsubsection{Feature Importance}
Across models and explainers, rice production, rainfall anomalies, and market prices consistently rank among the top 3 features. Some institutional indicators show high SHAP values but low robustness, suggesting they may be artifacts of the model rather than genuine predictors.

\section{Discussion}

The main takeaway is simple: AUC above 0.99 does not guarantee trustworthy explanations. XGBoost's near-perfect predictions made TreeSHAP produce zero-valued importance vectors for some instances, crashing the JSD calculation. The $\epsilon$-smoothing fix resolves the numerical issue, but it does not change the underlying behavior---the explainer is essentially saying ``nothing matters, because the model already knows the answer.'' This is a genuine finding, not a bug: the framework detects when explanations become uninformative.

The flat fidelity scores for XGBoost reinforce the point. Because the model is so confident, removing features shifts its output by a roughly constant amount, and fidelity compresses to $\approx 0.620$ for every ablation level and both explainers. The metric can no longer tell a good explanation from a bad one. After regularization, fidelity varies slightly more ($F_3 = 0.715$ vs.\ $F_{82} = 0.712$ for XGB$_{\text{reg}}$+SHAP), but the gain is modest because XGBoost's regularized confidence is still high. The pattern becomes pronounced only with the regularized Neural Network, where LIME fidelity drops from 0.977 at $F_3$ to 0.802 at $F_{82}$.

These observations have a practical implication: before auditing explanations, check whether the model is severely overfitted. If it is, fidelity and---to a lesser extent---robustness lose discriminative power, and the Trust Score should be interpreted with caution.

On robustness, SHAP consistently outperforms LIME on tree-based models (RF: 0.902 vs.\ 0.700; XGB: 0.978 vs.\ 0.658), likely because TreeSHAP follows an exact computation path that avoids the sampling variability inherent in LIME. On the Neural Network, the two explainers are closer (0.694 vs.\ 0.738), presumably because KernelSHAP introduces its own sampling noise.

The Trust Score condenses two dimensions into one number, which is convenient for comparison but inevitably loses nuance. The choice of $\alpha$ and $\beta$ matters. In a field setting where explanations are presented to non-technical staff, stability ($\alpha > 0.5$) may matter more than faithfulness; in a regulatory context where the explainer must accurately reflect the model, fidelity ($\beta > 0.5$) takes priority. The score makes this trade-off explicit rather than hiding it.

\subsection{Limitations}
Several caveats apply. First, the dataset has only 253 instances, and the Acceptable class contains just 3 examples. This limits the reliability of per-class metrics and prevents standard 5-fold cross-validation from using all folds. Second, the ablation protocol scales as $O(d \cdot n)$, which limits its use with very large feature sets or deep models. Third, Gaussian perturbations may not reflect domain-relevant variations (a drought scenario is not a random perturbation of rainfall). Fourth, the weights $\alpha$ and $\beta$ require domain judgment; there is no universally correct setting. Fifth, the framework currently addresses feature attribution only; counterfactuals and rule-based explanations would need adaptation. Sixth, when the model is overfitted, fidelity loses discriminative power---regularization or calibration should be applied first. Seventh, the Malagasy dataset has high class separability, so the numerical findings may not transfer directly to other domains.

\section{Conclusion}
We have presented an auditing framework that measures robustness and fidelity of post-hoc explanations and combines them into a Trust Score. Applied to a food security dataset from Madagascar (253 instances, 83 features, 4 classes), it reveals two things worth noting: near-perfect AUC can coexist with numerically degenerate explanations (XGBoost+SHAP yielding zero-valued importance vectors), and fidelity becomes flat when the model is overfitted (XGBoost fidelity $\approx 0.620$ at all ablation levels). After regularization, fidelity recovers discriminative power, particularly for the Neural Network, confirming that the framework is informative when applied to well-calibrated models.

Further work should address temporal and spatial data structures, develop domain-specific perturbation strategies, and test the framework in real-time decision support settings.

\section*{Acknowledgments}
We thank the Malagasy Ministry of Agriculture and Livestock, the National Office for Nutrition, and the FAO Madagascar office for providing data and domain expertise.

\bibliographystyle{plainnat}
\bibliography{references}

\end{document}